\documentclass{l4dc2026}
\title[ATOM-CBF]{ATOM-CBF: Adaptive Safe Perception-Based Control\\under Out-of-Distribution Measurements}

\usepackage{times}
\usepackage{amsmath}
\usepackage{amssymb}
\usepackage{algorithm}
\usepackage{algpseudocode}
\usepackage{hhline}
\usepackage{appendix}
\usepackage{rotating}
\usepackage{arydshln}
\usepackage{caption}
\usepackage{footmisc}
\usepackage{bm}
\usepackage{booktabs}
\usepackage{enumitem}
\usepackage{lipsum}
\usepackage{multirow}
\usepackage[table]{xcolor}
\usepackage{threeparttable}

\DeclareMathOperator*{\argmin}{arg\,min}

\newcommand\blfootnote[1]{%
  \begingroup
  \renewcommand\thefootnote{}\footnote{#1}%
  \addtocounter{footnote}{-1}%
  \endgroup
}

\definecolor{darkgreen}{rgb}{0.0, 0.5, 0.0}
\newcommand{\Unc}{\text{Unc}}
\newcommand{\EpsilonAdapt}{\epsilon_{\text{adapt}}}

\newtheorem{problem}{Problem}

\coltauthor{\Name{Kai S. Yun} \Email{kaisyun@mit.edu}\\
 \Name{Navid Azizan} \Email{azizan@mit.edu}\\
 \addr Massachusetts Institute of Technology, Cambridge, MA
}

\begin{document}

\maketitle
\thispagestyle{plain}

\begin{abstract}%
 Ensuring the safety of real-world systems is challenging, especially when they rely on learned perception modules to infer the system state from high-dimensional sensor data. These perception modules are vulnerable to epistemic uncertainty, often failing when encountering out-of-distribution (OoD) measurements not seen during training. To address this gap, we introduce ATOM-CBF (Adaptive-To-OoD-Measurement Control Barrier Function), a novel safe control framework that explicitly computes and adapts to the epistemic uncertainty from OoD measurements, without the need for ground-truth labels or information on distribution shifts. Our approach features two key components: (1) an OoD-aware adaptive perception error margin and (2) a safety filter that integrates this adaptive error margin, enabling the filter to adjust its conservatism in real-time. We provide empirical validation in simulations, demonstrating that ATOM-CBF maintains safety for an F1Tenth vehicle with LiDAR scans and a quadruped robot with RGB images.
\end{abstract}


\section{Introduction}
The problem of uncertainty lies at the heart of ensuring safety for autonomous systems in complex, real-world environments.
While a vast body of work provides safety guarantees for systems under uncertainties in their dynamics or parameters~\citep{xiao2021adaptive,lopez2023unmatched,siaGo2025yun}, this line of work often builds on a fundamental reliance on state information. 
In practice, states are not given in real-world deployment. 
Instead, they must be inferred from high-dimensional sensor measurements, such as camera images or LiDAR point clouds, often using learned perception modules, e.g., deep neural network (DNN)-based. 
These modules introduce their own critical source of uncertainty. 
To date, much of the field has focused on addressing safe perception-based control under \textit{aleatoric} uncertainty, e.g., stochastic noise inherent to sensors~\citep{Cosner2022SelfSupervisedOL,mrcbf-cp}. 
However, learned perception modules remain vulnerable to \textit{epistemic} uncertainty, i.e., the model's own lack of knowledge, which arises when encountering novel, out-of-distribution (OoD) data not seen during training. 

While epistemic uncertainty can be reduced by training the model on more diverse data~\citep{domain-randomization,hendrycks*2020augmix}, this approach is inherently limited, as it is untenable to capture the full distribution of all real-world scenarios. 
Frameworks that provide safety guarantees under OoD measurements often require new ground-truth labels to adapt online~\citep{antonio2024conformalPolicyLearning} or offline statistical bounds on a distribution of data~\citep{majumdar2021pacbayes}. These approaches are not designed for a truly on-the-fly safe controller, which has no access to ground-truth labels of a new OoD measurement and information on the distribution shift itself.
We address this gap by introducing \textbf{ATOM-CBF} (\textbf{A}daptive-\textbf{T}o-\textbf{O}oD-\textbf{M}easurement \textbf{C}ontrol \textbf{B}arrier \textbf{F}unction), a novel safe control framework that explicitly computes and adapts to the epistemic uncertainty from OoD measurements.


\begin{figure}[htp!]
    \centering
    \includegraphics[width=1.0\linewidth]{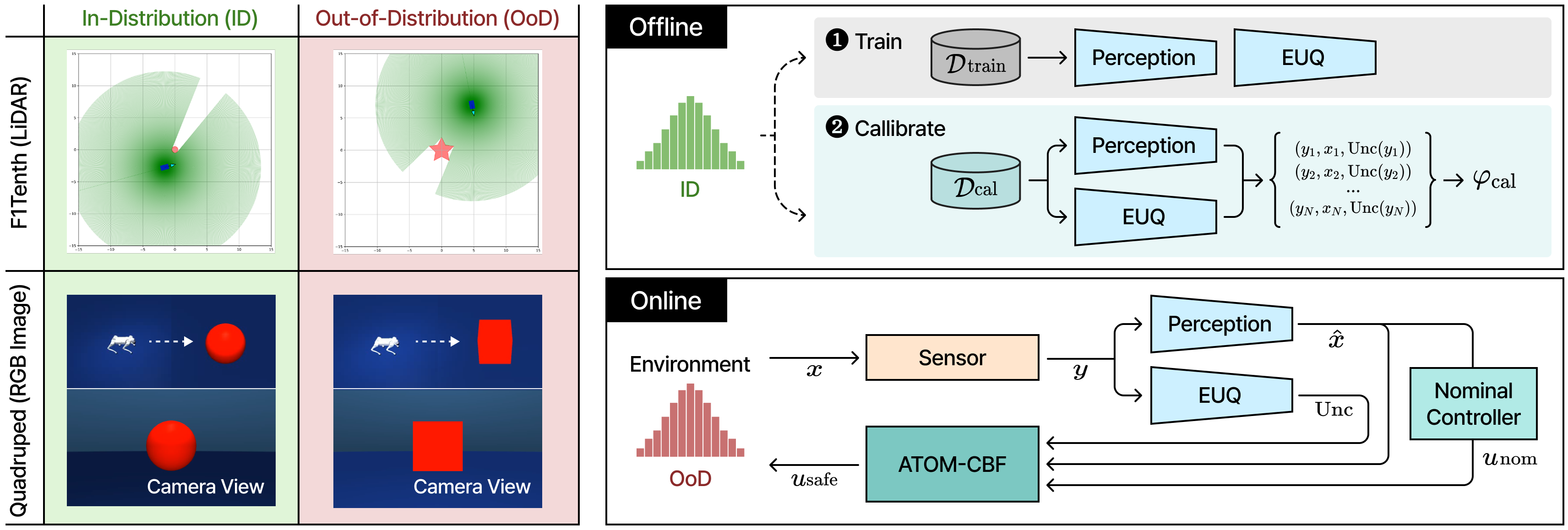}
    \caption{Problem setting (left) and ATOM-CBF (right). Offline, ID data is used to train a perception module and an epistemic uncertainty quantification (EUQ) module, and compute the base error ratio, $\varphi_{\text{cal}}$. Online, ATOM-CBF is deployed in OoD settings for F1Tenth and quadruped experiments.}
    \label{fig: main}
\end{figure}

\vspace{3mm}
\noindent\textbf{Contributions.}
In this work, we make the following contributions to provide empirical safety assurances for perception-based robotic control in the presence of OoD sensor measurements:
\vspace{1mm}
\begin{itemize}[itemsep=0.7ex, topsep=0pt, parsep=0pt, leftmargin=*]
    \item \textbf{OoD-aware adaptive perception error margin.} We introduce an effective calibration method to compute an adaptive error margin that dynamically scales with the epistemic uncertainty of a perception module (Fig.~\ref{fig: main}, Offline).
    \item \textbf{ATOM-CBF.} We propose the \textit{Adaptive-To-OoD-Measurement Control Barrier Function} (ATOM-CBF), a safety filter that integrates our adaptive error margin to ensure safety when encountering OoD sensor measurements (Fig.~\ref{fig: main}, Online).
    \item \textbf{Empirical Validation.} We demonstrate the efficacy of ATOM-CBF and provide empirical safety assurances in high-fidelity simulations, including an F1Tenth vehicle with 2D LiDAR scans and a quadruped robot with RGB camera images (Fig.~\ref{fig: main}, left panel).
\end{itemize}

\section{Related Work}
We group prior work on perception-based safety into two strands: (1) guaranteed safety under in-distribution (ID) measurements, i.e., data from the same distribution as the perception module's training set, and (2) OoD-aware safety.

\textbf{Safety guarantees for perception-based control (ID).} 
One line of work stem from formal verification, attempting to verify the perception DNN directly~\citep{brown2022unified,wei2025modelverification}, to be used with downstream safe controllers. However, these tools often struggle to scale to complex DNNs.
Another approach verifies the closed-loop system by abstracting the perception module.~\citet{mitra2022verify} use ``safe approximate abstractions,'' though the link to the real perception module is only empirical.~\citet{dean2021certaintyequiv} provide statistical guarantees but assume linear dynamics.~\citet{dawson2022certificate} present a certificate-based method that relies on an approximate sensor model, making it difficult to scale to high-dimensional sensor data, such as images.
A third category integrates perception uncertainty directly into the closed loop to provide robustness.~\citet{mrcbf} provide a control-theoretic formal bound of the worst-case perception error for robust safety. This has been extended to consider aleatoric uncertainty by~\citet{mrcbf-cp} using conformal prediction.~\citet{pwc} also use conformal prediction to provide robust bounding boxes around perceived obstacles, while other approaches numerically synthesize a safety function from perception data~\citep{pmlr-v242-toufighi24a,bena2025poisson}.
A key limitation across these methods is their dependency on bounds derived from ID data, making them vulnerable to OoD measurements.

\textbf{OoD-aware safety for perception-based systems.}
A second body of work addresses OoD-aware safety. Some methods seek to return the system to an ID state after an OoD detection.~\citet{richter2017safevisualnav} revert to safe prior behavior, and~\citet{reichlin2022backtomanifold} use a recovery policy to return to the ID manifold. 
Other approaches aim to prevent the system from ever entering OoD states.~\citet{wellhausen2020anomalyDetection} use anomaly detection to avoid OoD regions in a planner, while~\cite{castaneda2023idCBF},~\citet{seo2025unisafe}, and~\citet{bansal2024filterdeployallrobust} use barrier functions or reachability analysis to stay within ID regions.~\citet{chakraborty2024enhancingsafetyrobustnessvisionbased},~\citet{AgileButSafe}, and~\citet{contreras2025sodaMPC} instead use a fallback controller when OoD is detected.
However, these methods can fail if OoD entry and measurements are inevitable or the reaction is too late.
Finally, some frameworks provide safety guarantees under OoD measurements, but rely on information that may not always be available.~\citet{antonio2024conformalPolicyLearning} require new ground-truth labels online, and~\citet{majumdar2018pacbayes} require \textit{a priori} knowledge of the distribution shift bounds.

\textbf{Our work} instead provides empirical safety assurances under OoD measurements without requiring new ground-truth labels or \textit{a priori} distribution knowledge. We achieve this by introducing an OoD-aware adaptive perception error margin that scales in real-time with measured epistemic uncertainty, enabling our safety filter to dynamically adjust its conservatism.

\section{Preliminaries}
\blfootnote{\textbf{Notation.} $\mathbb R$ is the set of real numbers, $\mathbb R^n$ is a real vector, $\mathcal L$ is a Lie derivative, $\mathbb L$ is a Lipschitz constant, and $\|x\|_2$ is the euclidean norm for a vector $x$. $\kappa:\mathbb R\rightarrow\mathbb R$ denotes an extended class $\mathcal K_\infty$ function, i.e., a stictly increasing function where $\kappa(0)=0$, $\lim_{v\rightarrow-\infty}\kappa(v)=-\infty$, and $\lim_{v\rightarrow\infty}\kappa(v)=\infty$.}In this section, we provide a concise review of the concepts used to formulate our problem. We first introduce the system model and the notion of safety. Next, we review safe control under learned perception modules using Measurement-Robust CBF (MR-CBF), which provides safety under static, bounded perception errors. Finally, we discuss methods for epistemic uncertainty quantification (EUQ) of DNNs.

\subsection{System Model and Safety}
To start, consider a nonlinear control-affine system: 
\begin{align}\label{eq: sys_dyn}
    \dot x = f(x) + g(x)u,
\end{align}
where $x\in\mathbb R^n$ and $u\in\mathbb R^m$ are the state and control input, respectively, and functions $f:\mathbb R^n\rightarrow\mathbb R^n$ and $g:\mathbb R^n\rightarrow\mathbb R^{n\times m}$ are locally Lipschitz continuous. 
We define a safe set $\mathcal C\subset\mathbb R^n$ as the zero-superlevel set of a continuously differentiable function $h:\mathbb R^n\rightarrow\mathbb R$:
\begin{align}
    \mathcal C = \{ x\in\mathbb R^n: h(x)\ge0\}.
\end{align}
Safety of the system (\ref{eq: sys_dyn}) is achieved by ensuring that this set is control invariant, a widely used notion in the safe control literature~\citep{liu2014control,wei2019safe,ames2019control}. %
\begin{definition}\label{def: control_invariance}
    (Control Invariance). Let $x(t)$ denote the state trajectory of (\ref{eq: sys_dyn}) for time $t$ with initial state $x(0)$. The set $\mathcal C$ is control invariant if for every initial state $x(0)\in\mathcal C$, there exists an admissible control input $u$ such that the resulting state trajectory $x(t)\in\mathcal C,\;\forall t\ge 0$.
\end{definition}

\citet{ames2017cbf} introduces Control Barrier Function (CBF) as a method to formally guarantee this invariance for (\ref{eq: sys_dyn}) by providing a sufficient condition on the function $h(x)$. This condition ensures that for any state $x\in\mathcal C$, the set of admissible control inputs $u$ that render the system safe is non-empty.

\begin{definition}
    (Control Barrier Function (CBF)). Given a set $\mathcal C\subset\mathbb R^n$ defined as the zero-superlevel set of a continuously differentiable function $h:\mathbb R^n\rightarrow\mathbb R$, with 0 a regular value, $h$ is a control barrier function (CBF) for (\ref{eq: sys_dyn}) on $\mathcal C$ if there exists and extended class $\mathcal K_\infty$ function $\kappa$ such that
    \begin{equation}\label{eq: cbf}
        \sup_{u\in\mathcal U}\quad \underbrace{\frac{\partial h(x)}{\partial x}f(x)}_{\mathcal L_{f}h(x)} + \underbrace{\frac{\partial h(x)}{\partial x}g(x)}_{\mathcal L_{g}h(x)}u \ge -\kappa(h(x)).
    \end{equation}
\end{definition}

Note that the notion of safety depends on state information $x$. However, in many practical settings, such state information is not readily available and must be inferred from a state-dependent sensor measurement $y\in\mathbb R^l$. In our problem setting, we assume that this measurement, e.g., high-dimensional data such as LiDAR scans or camera images, is obtained via a locally Lipschitz continuous sensor map $p:\mathbb R^n\rightarrow\mathbb R^l$, such that $y = p(x)$.
We assume this relationship is deterministic, and note that the challenge of safe control using learned perception modules under stochastic sensor noise is addressed in~\citet{mrcbf-cp}.
A common assumption is the existence of a hypothetical inverse map $q:\mathbb R^l\rightarrow\mathbb R^n$ that can perfectly recover the state, i.e., $q(p(x))=x$. 

In practice, this map is often unknown. 
Thus, a learned perception map $\hat q:\mathbb R^l\rightarrow\mathbb R^n$ is used to approximate this ideal inverse map. In this work, we primarily consider deep neural network (DNN) perception maps $\hat q$ that are learned from data. Due to limitations in the learned model or deficiencies in the training data, the state estimate is related to the true state via an unknown error function $e:\mathbb R^n\rightarrow\mathbb R^n$,
\begin{align}\label{eq: perception_map}
    \hat x = \hat q(y) = x + e(x),
\end{align}
where the error function $e$ is implicitly defined by $\hat q$. This perception error $e(x)$ is the central challenge for perception-based safe control. CBF requires the true state $x$, but the controller only has access to the estimate $\hat x$. Applying a control input based on $\hat x$ without accounting for $e(x)$ can violate the safety constraint.

\subsection{Safe Control with Learned Perception Module}\label{subsec: mr_cbf}

To address this gap,~\citet{mrcbf} introduces measurement-robust control barrier function (MR-CBF). The MR-CBF framework thus provides a formal method to guarantee safety for systems relying on learned perception modules, provided the perception error is bounded and known.
\begin{definition}
    (Measurement-Robust Control Barrier Function (MR-CBF)). Let $\mathcal C\subset\mathbb R^n$ be the zero-superlevel set of a continuously differentiable function $h:\mathbb R^n\rightarrow\mathbb R$ with 0 a regular value. Then, the function $h$ is a measurement-robust control barrier function (MR-CBF) for system (\ref{eq: sys_dyn}) on $\mathcal{C}$ with parameter function pair $(a,b):\mathbb R^l \rightarrow \mathbb R_+^2$ if there exists an extended class $\mathcal K_\infty$ function $\kappa$ such that
    \begin{equation}\label{eq: mr_cbf}
        \sup_{u\in\mathbb R^m} \bigg\{ \mathcal L_fh(\hat x) + \mathcal L_gh(\hat x)u - \big(a(y) + b(y)\|u\|_2\big) \bigg\} \ge -\kappa(h(\hat x)).
    \end{equation}
\end{definition}
The terms $a(y)$ and $b(y)$ in (\ref{eq: mr_cbf}) create a measurement-dependent ``robustness buffer,'' forcing the controller to be more conservative to account for the perception error. The key result in~\citet{mrcbf} connects these abstract parameters to a concrete perception error bound $\epsilon(y)$, which is static. If the perception error is bounded such that $\|e(x)\|_2 \le \epsilon(y)$ for all $x \in \mathcal{C}$, and the functions $\mathcal L_f h$, $\mathcal L_g h$, and $\kappa \circ h$ are Lipschitz continuous on $\mathcal C$ with Lipschitz coefficients $\mathbb{L}_{\mathcal L_f h}$, $\mathbb{L}_{\mathcal L_g h}$, and $\mathbb{L}_{\kappa \circ h}$, respectively, then safety is guaranteed by setting $a(y) = \epsilon(y)(\mathbb{L}_{\mathcal L_f h} + \mathbb{L}_{\kappa \circ h})$ and $b(y) = \epsilon(y)\mathbb{L}_{\mathcal L_g h}$. 

\begin{remark}\label{remark: mr_cbf_limit}
    The MR-CBF framework provides robustness by assuming a known, static error bound $\epsilon(y)$. This approach is effective for systems that deal with ID data or for handling aleatoric uncertainty, i.e., sensor noise. However, it does not account for epistemic uncertainty. This becomes critical when a perception module encounters novel OoD measurements, as the true error can far exceed the assumed $\epsilon(y)$, leading to safety violations. Furthermore, the validation of these formal error bounds in prior work has centered on classical models like Kernel Ridge Regression or constant offsets, not the high-dimensional DNNs that are highly susceptible to such OoD failures.
\end{remark}

\subsection{Epistemic Uncertainty Quantification for Learned Perception Modules}

To address the gap identified in Remark~\ref{remark: mr_cbf_limit}, the downstream safe controller must be able to account for epistemic uncertainty of OoD measurements in real-time. This requires an associated epistemic uncertainty quantification (EUQ) module, $\text{Unc}:\mathbb R^l\rightarrow\mathbb R_+$. This module must produce a scalar score $\text{Unc}(y)$ that is low for ID measurements and high for OoD measurements. 
We focus on two prominent EUQ modules that represent a key trade-off: Deep Ensembles as a high-performance module, and SCOD as a computationally-efficient, post-hoc alternative. See Appendix~\ref{app: euq} for details.

\textbf{Deep Ensembles.}~\citet{deep_ensemble} introduce a simple and high-performing non-Bayesian approach for epistemic uncertainty quantification. Deep Ensemble involves training an ensemble of $M$ networks, e.g., $M=5$, with identical architectures but different random initializations. At runtime, all $M$ networks make a prediction. The epistemic uncertainty is then quantified by the variance in their outputs. If the models disagree, the uncertainty is high.
While this method is effective at producing high-quality epistemic uncertainty estimates for OoD inputs, it requires the training and inference of $M$-number of networks, making it a computationally expensive procedure.

\textbf{SCOD.}~\citet{SCOD} propose Sketching Curvature for OoD Detection (SCOD), a model architecture-agnostic method that equips a single, pre-trained network with an uncertainty score post-hoc. It builds on the Laplace approximation~\citep{MacKay1992APB}, which uses the curvature of the loss landscape (characterized by the Fisher information matrix) to estimate epistemic uncertainty. SCOD operates in two phases. Offline, it computes and stores a tractable, low-rank approximation (a ``sketch") of the training data's Fisher matrix. Online, it compares a new input's local curvature to the known curvature of the training data. A significant mismatch results in a high uncertainty score. Unlike Deep Ensemble, SCOD only needs a single pre-trained model, making it an efficient option for real-time deployment.~\citet{SCOD} show that SCOD's OoD-detection ability often matches or exceeds Deep Ensemble's, with favorable runtime/AUROC Pareto efficiency. 

\section{ATOM-CBF: Adaptive-To-OoD-Measurement Control Barrier Function}
The concepts reviewed thus far highlight our central challenge. We aim to design a controller that ensures safety in the sense of control invariance (Def.~\ref{def: control_invariance}) for systems relying on an imperfect DNN-based perception module $\hat q$ in the presence of OoD measurements. The MR-CBF framework (Sec.~\ref{subsec: mr_cbf}) provides a path, but its reliance on a known, static error bound $\epsilon(y)$ makes it vulnerable to OoD data, where unmodeled epistemic uncertainty $\Unc(y)$ can cause the true perception error to violate this bound. This motivates our problem: \textit{to design a safe control framework that adapts its robustness by explicitly incorporating the measured epistemic uncertainty, without access to ground-truth states or any information about the OoD distribution}. We formalize this as follows:

\begin{problem}\label{prob_1}
    Consider the system dynamics (\ref{eq: sys_dyn}) with an initial state $x(0)\in\mathcal C$, sensor map $p$, and learned perception map $\hat q$ in (\ref{eq: perception_map}). Let $h:\mathbb R^n\rightarrow\mathbb R$ be a continuously differentiable constraint function defining the safe set $\mathcal C$, and let $\Unc:\mathbb R^l\rightarrow\mathbb R_+$ be an EUQ module that provides an estimate of the epistemic uncertainty for a given measurement $y$. Design a safe control framework, consisting of: (1) an adaptive perception error margin, $\epsilon_{\text{adapt}}(y)$, as a function of the epistemic uncertainty score $\Unc(y)$, and (2) a safe control law $u_{\text{safe}}=k(\hat x, \epsilon_{\text{adapt}}(y))$ that generates an admissible input $u\in u_{\text{safe}}$ such that for every $x(0)\in\mathcal C$, $x(t)\in\mathcal C,\forall t\ge0$.
\end{problem}

\subsection{OoD-Aware Adaptive Perception Error Margin}\label{subsec: adapt_bound}

This section details the process for computing our \textit{OoD-aware adaptive perception error margin}. This bound is a critical component that will enable our downstream safety filter to adjust its conservatism in real-time, providing adaptive safety even when encountering OoD measurements.

\begin{enumerate}[itemsep=1.0ex, topsep=0pt, parsep=0pt, leftmargin=*]%
    \item \textbf{Filtered Calibration Set}. We start with an initial calibration dataset, $\mathcal D_{\text{cal}}=\{(y_i, x_i)\}_{i=1}^N$, where $y_i$ is a measurement, $x_i$ is the corresponding ground-truth state, and $\mathcal D_{\text{cal}}$ is drawn from the identical distribution as the training dataset. To create a stable set, we first compute the uncertainty scores for the calibration data, $S_{\text{cal}} = \{ \text{Unc}(y_i) \}_{i=1}^N$, and compute its mean $\mu_{\text{unc}}$. We then introduce a user-defined filter hyperparameter $\gamma >0$, which sets an absolute tolerance around the mean. The \textit{filtered ID set}, $\mathcal D_{\text{filtered}}$, is then defined by removing statistical outliers:
    \begin{align}\label{eq: unc_filtered_set}
        \mathcal D_{\text{filtered}} \triangleq \left\{ (y_i,x_i)\in\mathcal D_{\text{cal}} : |\text{Unc}(y_i) - \mu_{\text{unc}}|\le\gamma \right\}.
    \end{align}

    \item \textbf{Base Error Ratio}. We define the \textit{base error ratio}, $\varphi_{\text{cal}}\in\mathbb R^n_+$, where each $j$-th element represents the worst-case ratio of the element-wise true estimation error to the measured epistemic uncertainty, computed over the filtered ID set:
    \begin{align}\label{eq: base_error_ratio}
        \varphi_{\text{cal},j} \triangleq \max_{(y_i,x_i)\in\mathcal D_{\text{filtered}}} \left( \frac{| \hat q_j(y_i)-x_{i,j}|}{\text{Unc}(y_i)} \right), \quad \forall j\in \{1,\ldots,n\},
    \end{align} 
    where $\hat q_j(y_i)$ and $x_{i,j}$ are the $j$-th components of the estimated and true state vectors, respectively. This definition of $\varphi_{\text{cal}}$ is inspired by the method of conformalizing scalar uncertainty estimates~\citep{cp_bates_intro}, where we effectively set the risk level to be the worst-case. See Appendix~\ref{app: base_error_ratio} for details.
    \item \textbf{Adaptive Perception Error Margin}. For any new measurement $y$ encountered during deployment, we compute its epistemic uncertainty $\text{Unc}(y)$. Using this, we define the adaptive perception error margin, $\epsilon_{\text{adapt}}(y):\mathbb R^l\rightarrow\mathbb R_+$:
    \begin{align}\label{eq: adaptive_error_bound}
        \epsilon_{\text{adapt}}(y) \triangleq \left\|\varphi_{\text{cal}} \cdot \text{Unc}(y)\right\|_2.
    \end{align}
\end{enumerate}
The adaptive perception error margin, $\epsilon_{\text{adapt}}$ is the core mechanism of our method. If the measurement $y$ is OoD, the EUQ module will output a high uncertainty score, $\text{Unc}(y)$. This dynamically and proportionally increases the assumed error bound $\epsilon_{\text{adapt}}$, forcing the downstream safety filter to be more conservative to account for the high uncertainty of the OoD measurement. 

\begin{remark}
    The filtering procedure in (\ref{eq: unc_filtered_set}) is a stabilization heuristic to prevent $\varphi_{\text{cal}}$ from being skewed by outliers, i.e., data points with anomalously low uncertainty ($\Unc(y_i)\approx 0$) but non-trivial perception error. This step prevents the base error ratio (\ref{eq: base_error_ratio}) from becoming arbitrarily large and rendering the downstream safety filter overly conservative.
    Sec.~\ref{subsec: exp_2d} demonstrates this empirically.
\end{remark}

\subsection{Adaptive Safe Control against OoD Measurements}
Having defined the OoD-aware adaptive perception error margin $\EpsilonAdapt(y)$ in Sec.~\ref{subsec: adapt_bound}, we now address the second part of Problem~\ref{prob_1}: constructing the safe control law. Our approach is to integrate this adaptive margin directly into the MR-CBF framework as follows.

\begin{definition}\label{def: atom_cbf}
    (\textbf{A}daptive-\textbf{T}o-\textbf{O}oD-\textbf{M}easurement \textbf{C}ontrol \textbf{B}arrier \textbf{F}unction (ATOM-CBF)). Let $\mathcal C\subset\mathbb R^n$ be the zero-superlevel set of a continuously differentiable function $h:\mathbb R^n\rightarrow\mathbb R$. Let $\epsilon_{\text{adapt}}(y) = \left\| \varphi_{\text{cal}} \cdot \text{Unc}(y)\right\|_2$ be the OoD-aware adaptive perception error margin, given an OoD measurement $y$. Here, $\varphi_{\text{cal}}$ is the base error ratio from (\ref{eq: base_error_ratio}) and $\text{Unc}(y)$ is the output of an epistemic uncertainty quantification (EUQ) module. The function $h$ is an Adaptive-to-OoD-Measurement Control Barrier Function (ATOM-CBF) for system (\ref{eq: sys_dyn}) on $\mathcal C$ if there exists an extended class $\mathcal K_{\infty}$ function $\kappa$ such that
    \begin{equation}\label{eq: atom_cbf}
        \sup_{u\in\mathbb R^m} \bigg\{ \mathcal L_fh(\hat x) + \mathcal L_gh(\hat x)u - \epsilon_{\text{\normalfont adapt}}(y) \big(\mathbb L_{\mathcal L_fh} + \mathbb L_{\kappa\circ h} + \mathbb L_{\mathcal L_gh}\|u\|_2\big) \bigg\} \ge -\kappa(h(\hat x)).
    \end{equation}
\end{definition}

Compared to (\ref{eq: mr_cbf}), (\ref{eq: atom_cbf}) does not have the robustness terms $a(y)$ and $b(y)$ fixed with a static $\epsilon(y)$. 
Instead, they adapt based on the online estimate of the epistemic uncertainty, providing an adaptive robustness buffer that scales with the measured epistemic uncertainty.
We now introduce our optimization-based safety filter as follows:
\begin{align}\label{eq: atom_op}
    u_{\text{safe}}(\hat x, \epsilon_{\text{adapt}}(y)) = &\argmin_{u\in\mathbb R^m} \quad \frac12 \| u - u_{\text{nom}}\|_2^2 \\
    \text{s.t.}\quad&\ \mathcal L_fh(\hat x) + \mathcal L_gh(\hat x)u - \epsilon_{\text{adapt}}(y) \big(\mathbb L_{\mathcal L_fh} + \mathbb L_{\kappa\circ h} + \mathbb L_{\mathcal L_gh}\|u\|_2\big) \ge -\kappa(h(\hat x)), \nonumber 
\end{align}
where $u_{\text{nom}}$ is a potentially unsafe control input from a nominal controller, e.g. goal-reaching controller.
The constraint in (\ref{eq: atom_op}) is non-smooth, making the optimization problem a second-order cone program (SOCP). To ensure constraint feasibility in practice, we follow the approach in~\citet{mrcbf} and introduce a slack variable, $\delta$, to the CBF constraint, which is then heavily penalized in the cost function. See Appendix~\ref{app: ood} for implementation details. 

\section{Experimental Results}
We test ATOM-CBF on two test beds: a 2-dimensional F1Tenth vehicle with LiDAR scans (Sec.~\ref{subsec: exp_2d}) and a 3-dimensional quadruped robot with RGB camera images (Sec.~\ref{subsec: exp_3d}).
In both experiments, the perception modules are DNNs trained to detect a single, static obstacle. 
While the experimental details differ, both experiments share a common foundation.

\textbf{\underline{System Dynamics}}. We model both systems using the 2D unicycle dynamics with respect to a static obstacle:
\begin{equation}\label{eq: uni_local_dyn}
    \dot x = \begin{bmatrix}
        \dot d\\ \dot\alpha
    \end{bmatrix} = \underbrace{\mathbf{0}_{2\times 1}}_{f(x)} + \underbrace{\begin{bmatrix}
        -\cos(\alpha) & 0 \\ -\sin(\alpha)/d & 1
    \end{bmatrix}}_{g(x)} \underbrace{\begin{bmatrix}
        v\\\omega
    \end{bmatrix}}_{u},
\end{equation}
\noindent
\begin{minipage}{0.19\textwidth}
    \centering
    \includegraphics[width=0.90\linewidth]{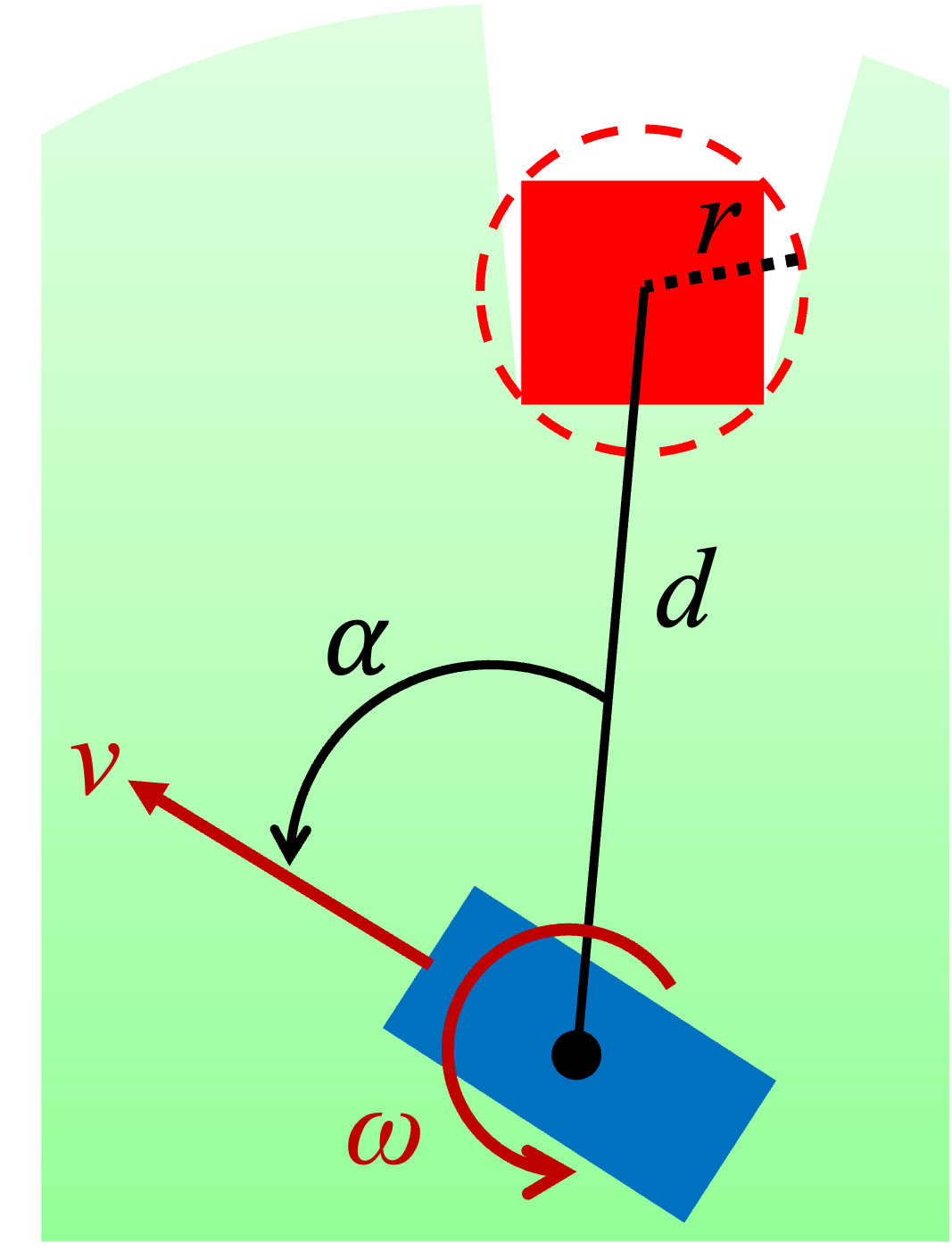}
    \captionof{figure}{F1/10.}
    \label{fig: f1tenth_sys}
\end{minipage}\hfill
\begin{minipage}{0.79\textwidth}
    where the state $x=[d,\alpha]^\top$ consists of distance $d$ to a static obstacle and relative heading angle $\alpha$ with respect to the obstacle. The control inputs are the longitudinal velocity $v$ and yaw rate $\omega$. 
    This reduced-order model (ROM) abstracts away the low-level control of the robot to focus on high-level navigation. 
    Both experiments consider various shapes of obstacles: F1Tenth experiment uses circle, triangle, square, and star obstacles, and quadruped experiment uses sphere and cube obstacles.
    The unsafe region is defined by the minimum enclosing circle of radius $r$ of the obstacle in the 2-dimensional navigation space, irrespective of the obstacle geometry, as shown in Fig.~\ref{fig: f1tenth_sys}. 
\end{minipage}
\\
\\
\indent\textbf{\underline{Calibration and Base Error Ratio}}. The calibration procedure from Sec.~\ref{subsec: adapt_bound} with $\gamma=\sigma_{\text{unc}}$, i.e., standard deviation of $S_{\text{cal}}$, is used for both experiments, though with different $\mathcal D_{\text{filtered}}$ datasets. Full statistics are in Appendix~\ref{app: ood}.

\textbf{\underline{Safety Objective and Filters}}. For both experiments, the safety objective is to avoid collision with the obstacle. 
For this, we employ a state-based cone CBF, inspired by Collision Cone CBFs~\citep{thontepu2022control,tayal2023control,tayal2024collisionconeapproachcontrol}. This CBF ensures that the vehicle does not point toward any detected obstacle: $h(\hat x)=|\hat\alpha| - \sin^{-1}(r/\hat d)$, where the radius $r$ is assumed to be known and $\hat d > r$ is enforced for all timesteps. 
Using this CBF, we construct and compare the following three safety filters:

\begin{enumerate}[itemsep=1.0ex, topsep=0pt, parsep=0pt, leftmargin=*]
    \item \textbf{CBF-QP}: Baseline; a standard CBF-QP with constraint (\ref{eq: cbf}), as in~\citet{ames2017cbf}.
    \item \textbf{ATOM-CBF (SCOD)}: Our SOCP adaptive filter (\ref{eq: atom_op}), with SCOD as the EUQ module.
    \item \textbf{ATOM-CBF (Deep)}: Our SOCP adaptive filter (\ref{eq: atom_op}), with Deep Ensemble as the EUQ module.
\end{enumerate}
Note that for each experiment, all safety filters receive their state estimate $\hat x=[\hat d,\hat\alpha]^\top$ from the identical perception module trained on in-distribution (ID) data.

\subsection{2D Experiment: F1Tenth Vehicle Control with LiDAR Scans}\label{subsec: exp_2d}

\begin{table}[H]
    \centering
    {\small
    \begin{threeparttable}
    \caption{Simulation results for F1Tenth vehicle control (1,000 trajectories each).\tnote{a}}
    \label{tab: sim_f1}
    \begin{tabular}{lcccc} 
        \toprule
         & \textbf{ID (Circle)} & \multicolumn{3}{c}{\textbf{OoD (Square, Triangle, Star)}} \\
        \cmidrule(lr){2-2} \cmidrule(lr){3-5}
        \textbf{Safety Filter}    & CBF-QP & CBF-QP & \textbf{ATOM-CBF (SCOD)} & \textbf{ATOM-CBF (Deep)} \\
        \midrule
        \rowcolor{blue!20}
        Reach        & 100\%       & 63.60\%      & 96.10\%     & 33.70\% \\
        \rowcolor{green!20}
        Deadlock     & 0.00\%      & 18.40\%      & 3.90\%               & 66.30\% \\
        \rowcolor{red!20}
        Collision    & 0.00\%      & 18.00\% & 0.00\%               & 0.00\% \\
        \midrule
        $d$-Coverage   & ---         & ---          & 20.58\%              & 70.27\% \\
        $\alpha$-Coverage & ---      & ---          & 54.35\%              & 99.99\% \\
        \midrule
        AUROC        & ---         & ---          & 0.9563               & 0.9971 \\
        \bottomrule
    \end{tabular}
    \begin{tablenotes}
            \item[a] See Appendix~\ref{app: add_exp_details} for details on calculations for coverages and AUROC.
        \end{tablenotes}
    \end{threeparttable}
    }
\end{table}

\begin{figure}[ht!]
    \captionsetup{skip=1pt}
    \centering
    \includegraphics[width=0.95\linewidth]{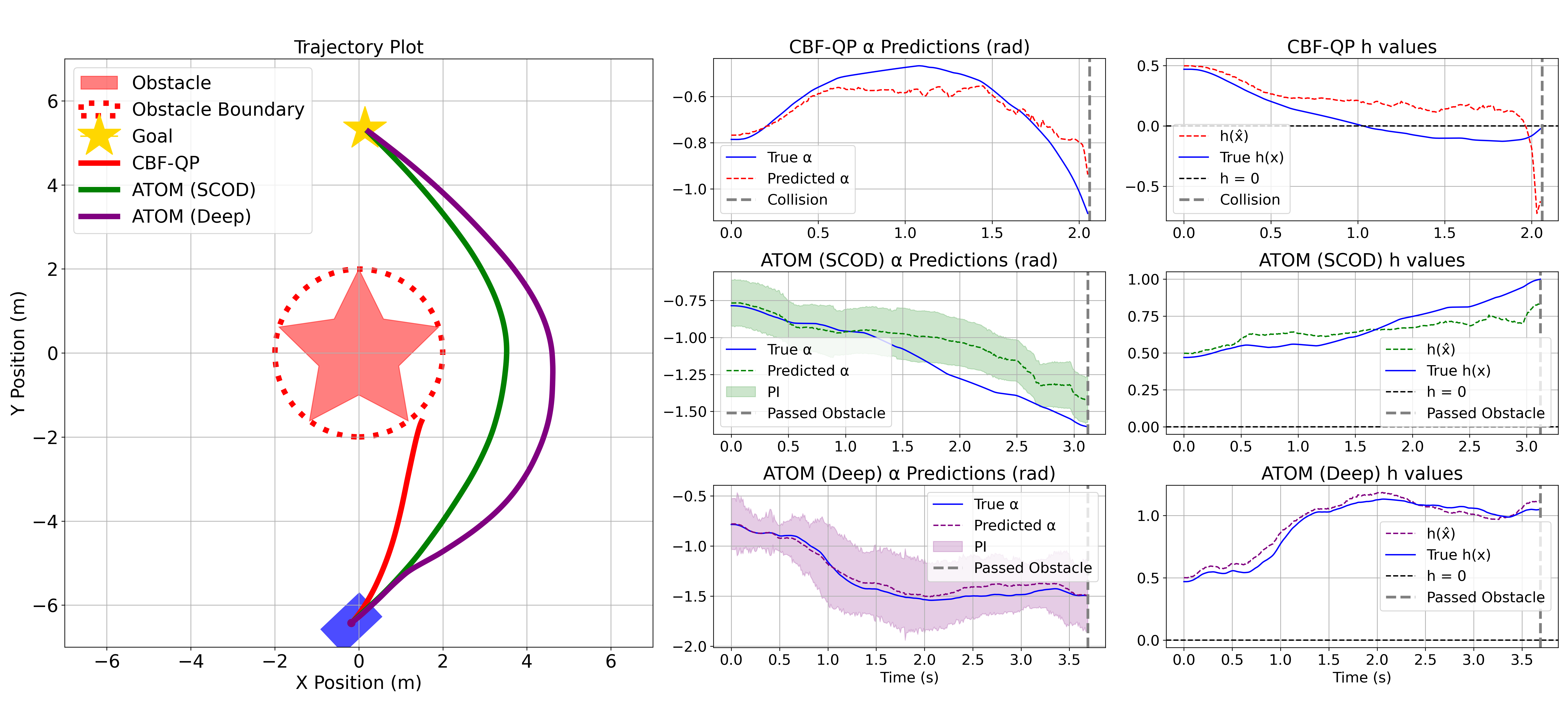}
    \caption{F1Tenth vehicle experiment with a star obstacle. All trajectories start with an identical initial condition and OoD obstacle, comparing three controller variants: CBF-QP (red), ATOM-CBF with SCOD (green), ATOM-CBF with Deep Ensemble (purple). (Left) Trajectory plot. (Middle) Time plots of true $\alpha$ (blue) vs. predicted $\hat\alpha$ and its prediction interval (PI). (Right) Time plots of true $h(x)$ (blue) vs. estimated $h(\hat x)$. Perception and safety filters are engaged until the vehicle passes the obstacle, at which point the nominal controller resumes control using the ground-truth state.}
    \label{fig: sim_f1}
\end{figure}

\textbf{\underline{Environment and Perception}}. We employ the F1Tenth Gym platform~\citep{f1tenth}, where the sensor map $p$ models a 1,080-dimensional 2D LiDAR scan with $360^\circ$ field-of-view (FOV). The perception map is a convolutional neural network (CNN) that takes this 1,080-dimensional LiDAR measurement and outputs the estimated state, $[\hat d, \hat\alpha]^\top$. 

\textbf{\underline{Nominal Objective}}. The nominal PD controller directs the vehicle straight to a goal behind the obstacle using perfect global state information, $[x,y,\theta]^\top$. This forces the safety filter to intervene.

\textbf{\underline{ID vs. OoD}}. The perception module was trained on ID data of small (0.1 - 0.5 m radius) \textit{circle} obstacles. The OoD challenge introduces geometrically distinct (\textit{square, triangle, star}) and much larger (1.5 - 2.0 m side-length) obstacles. See Fig.~\ref{fig: main} for the significant difference in obstacles.

\textbf{\underline{Results and Analysis}}. Table~\ref{tab: sim_f1} summarizes the results.
The CBF-QP baseline, while 100\% successful on ID data, collided in 18\% of OoD trials.
In contrast, both ATOM-CBF variants achieve a 0\% collision rate, successfully adapting to OoD measurements. 
However, Table~\ref{tab: sim_f1} reveals a key performance trade-off. ATOM-CBF (Deep) is overly conservative, leading to a high deadlock rate (66.30\%), whereas ATOM-CBF (SCOD) achieves safety with a 96.10\% reach rate.
This stems from Deep Ensemble's large $\Unc(y)$ jump for OoD data compared to ID data, producing a higher AUROC, larger $\EpsilonAdapt(y)$, and greater robustness buffer.
This over-conservatism is reflected in its 99.99\% $\alpha$-coverage and wide PI in Fig.~\ref{fig: sim_f1}. While ATOM-CBF (Deep) trajectory (purple) takes a much wider path, ATOM-CBF (SCOD) trajectory (green) is tighter with a smaller buffer.
Note that high coverage is not our goal, but we include it to illustrate the large, conservative bounds that cause the safety filter to deadlock. See Appendix~\ref{app: ood} for details on $\Unc$ scores in OoD compared to ID.

The success of ATOM-CBF (SCOD) hinges on the filtering heuristic (Sec.~\ref{subsec: adapt_bound}), which is necessary to filter statistical outliers that cause the base error ratio, $\varphi_{\text{cal}}$, to become unreasonably large. Table~\ref{tab: abl_f1} validates this by ablating the filter width $\gamma$. 
As $\gamma$ increases to $5\cdot\sigma_{\text{unc}}$, it includes these outliers, causing $\varphi_{\text{cal}}$ to jump to the unfiltered value, i.e., when $\gamma=\infty$. This increased $\varphi_{\text{cal}}$ results in over-conservatism, causing performance to drop drastically. This shows $\gamma$ is a tunable parameter for adjusting the desired conservatism. Our chosen value, $\gamma=\sigma_{\text{unc}}$, proves to be an effective heuristic.

\begin{table}[htp!]
    \centering
    \caption{Ablation study on the hyperparameter $\gamma$ for ATOM-CBF (SCOD) (1,000 trajectories each).}
    {\small
    \begin{tabular}{cc>{\columncolor{blue!20}}c>{\columncolor{green!20}}cccc}
        \toprule
        \multicolumn{2}{c}{\textbf{SCOD: Filtering \& Calibration}} & \multicolumn{5}{c}{\textbf{ATOM-CBF (SCOD): F1Tenth Result}}\\
        \cmidrule(lr){1-2} \cmidrule(lr){3-7}
        {\boldmath$\gamma$} & $\varphi_{\text{cal}}([d,\alpha])$ & Reach & Deadlock & Collision & $d$-coverage & $\alpha$-coverage \\
        \midrule
        {\boldmath$\sigma_{\text{unc}}$} & [3.690e-2, 1.777e-2] & 96.10\% & 3.90\% & 0.00\% & 20.58\% & 54.35\%\\
        {\boldmath$2\cdot\sigma_{\text{unc}}$} & [3.959e-2, 1.963e-2] & 92.50\% & 7.50\% & 0.00\% & 24.51\% & 51.68\%\\
        {\boldmath$4\cdot\sigma_{\text{unc}}$} & [3.961e-2, 1.963e-2] & 90.30\% & 9.70\% & 0.00\% & 26.81\% & 56.48\%\\
        {\boldmath$5\cdot\sigma_{\text{unc}}$} ($\approx\infty$) & [8.574e-2, 2.260e-2] & 28.30\% & 71.70\% & 0.00\% & 57.25\% & 79.75\%\\
        \bottomrule
    \end{tabular}
    }
    \label{tab: abl_f1}
\end{table}
\vspace{-1em}

\subsection{3D Experiment: Quadruped Control with RGB Camera Images}\label{subsec: exp_3d}

\textbf{\underline{Environment and Perception}}. Here, we use a Unitree Go2 robot in a 3D MuJoCo environment. The sensor map $p$ models a $1280\times 720$ ($58^\circ$ vertical FOV) RGB feed from the quadruped's fixed-height head camera. The perception map is a CNN that outputs estimated state, $[\hat d, \hat\alpha]^\top$. 

\textbf{\underline{Nominal Objective}}. The same nominal PD controller now acts as an adversarial, directing the quadruped to cause a collision, as shown in Fig.~\ref{fig: quad_crash}. In this experiment, there is no goal point to reach.
\\
\\
\noindent
\begin{minipage}{0.35\textwidth}
    \centering
    \includegraphics[width=0.90\linewidth]{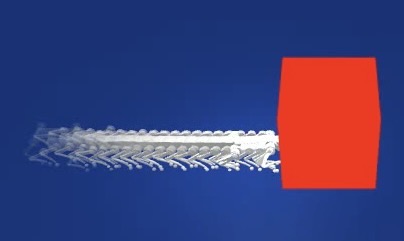}
    \captionof{figure}{Quadruped crash (OoD).}
    \label{fig: quad_crash}
\end{minipage}\hfill
\begin{minipage}{0.63\textwidth}
    \textbf{\underline{ID vs. OoD}}. The perception module was trained on ID data of high-contrast scenes with a sphere obstacle. The OoD challenge simulates dense fog by using high ambient and low diffuse light, creating a ``washed-out'' effect, i.e., a low-contrast environment where the obstacle and background are difficult to distinguish. Furthermore, the obstacle is a cube with side length 1.0 m. See Fig.~\ref{fig: main} for the visual difference in ID vs. OoD scenery. Fig.~\ref{fig: quad_crash} shows CBF-QP crashing in OoD.
\end{minipage}
\\

\textbf{\underline{Results and Analysis}}. 
We evaluated each safety filter over 100 trials. 
In the ID setting (high-contrast, sphere), the baseline CBF-QP collided in only 2\% of trials. However, in the OoD challenge (low-contrast, cube), the CBF-QP collided in 97\% of trajectories. In stark contrast, both ATOM-CBF variants achieved a 0\% collision rate in the OoD setting, demonstrating their ability to adapt to the challenging measurements and maintain safety. AUROC is above 0.99 for both EUQs.

\section{Conclusion and Future Work}
In this work, we propose ATOM-CBF, a novel framework for safe perception-based control under OoD measurements. It features an OoD-aware adaptive perception error margin and a safety filter that integrates this margin, allowing the controller's conservatism to scale with real-time epistemic uncertainty. We empirically validated ATOM-CBF in two distinct high-fidelity robotic environments, demonstrating that it maintains safety in challenging OoD scenarios.
The efficacy of our approach is linked to the performance of the underlying EUQ module, and the safety assurances are contingent on the EUQ module's capacity to reliably detect novel OoD inputs. Our experiments further highlight a trade-off: the choice of EUQ module and filtering hyperparameter dictates the controller's conservatism and task performance. Future work includes developing methods to further refine the safety-performance trade-off and deploying ATOM-CBF on real-world hardware.

\acks{The authors acknowledge the MIT SuperCloud and Lincoln Laboratory Supercomputing Center for providing computing resources that have contributed to the results reported within this paper. The authors thank Zeyang Li (MIT) for valuable discussions, Lars Lindemann (ETH Z\"{u}rich) and Shuo Yang (formerly UPenn) for sharing related code, and Suyoung Kwon (KAIST) for help with illustrations.}

\bibliography{main}

@misc{cp_bates_intro,
      title={A Gentle Introduction to Conformal Prediction and Distribution-Free Uncertainty Quantification}, 
      author={Anastasios N. Angelopoulos and Stephen Bates},
      year={2022},
      eprint={2107.07511},
      archivePrefix={arXiv},
}

@InProceedings{SCOD,
  title = 	 {Sketching curvature for efficient out-of-distribution detection for deep neural networks},
  author =       {Sharma, Apoorva and Azizan, Navid and Pavone, Marco},
  booktitle = 	 {Proceedings of the Thirty-Seventh Conference on Uncertainty in Artificial Intelligence},
  pages = 	 {1958--1967},
  year = 	 {2021},
  series = 	 {Proceedings of Machine Learning Research},
  publisher =    {PMLR},
}

@InProceedings{mrcbf,
  title = 	 {Guaranteeing Safety of Learned Perception Modules via Measurement-Robust Control Barrier Functions},
  author =       {Dean, Sarah and Taylor, Andrew and Cosner, Ryan and Recht, Benjamin and Ames, Aaron},
  booktitle = 	 {Proceedings of the 2020 Conference on Robot Learning},
  year = 	 {2021},
  series = 	 {Proceedings of Machine Learning Research},
  publisher =    {PMLR},
}

@INPROCEEDINGS{mrcbf-cp,
  author={Yang, Shuo and Pappas, George J. and Mangharam, Rahul and Lindemann, Lars},
  booktitle={2023 62nd IEEE Conference on Decision and Control (CDC)}, 
  title={Safe Perception-Based Control Under Stochastic Sensor Uncertainty Using Conformal Prediction}, 
  year={2023},
}

@InProceedings{pwc,
  title = 	 {Perceive With Confidence: Statistical Safety Assurances for Navigation with Learning-Based Perception},
  author =       {Dixit, Anushri and Mei, Zhiting and Booker, Meghan and Storey-Matsutani, Mariko and Ren, Allen Z. and Majumdar, Anirudha},
  booktitle = 	 {Proceedings of The 8th Conference on Robot Learning},
  pages = 	 {2517--2541},
  year = 	 {2025},
  series = 	 {Proceedings of Machine Learning Research},
  publisher =    {PMLR},
}

@inproceedings{deep_ensemble,
  title={Simple and Scalable Predictive Uncertainty Estimation using Deep Ensembles},
  author={Balaji Lakshminarayanan and Alexander Pritzel and Charles Blundell},
  booktitle={Advances in Neural Information Processing Systems},
  year={2017},
  pages={6402-6413}
}

@inproceedings{f1tenth,
  title={F1TENTH: An Open-source Evaluation Environment for Continuous Control and Reinforcement Learning},
  author={O’Kelly, Matthew and Zheng, Hongrui and Karthik, Dhruv and Mangharam, Rahul},
  booktitle={NeurIPS 2019 Competition and Demonstration Track},
  pages={77--89},
  year={2020},
  organization={PMLR}
}

@inproceedings{wei2019safe,
  title={Safe control algorithms using energy functions: A unified framework, benchmark, and new directions},
  author={Wei, Tianhao and Liu, Changliu},
  booktitle={2019 IEEE 58th Conference on Decision and Control (CDC)},
  pages={238--243},
  year={2019},
  organization={IEEE}
}

@inproceedings{liu2014control,
  title={Control in a safe set: Addressing safety in human-robot interactions},
  author={Liu, Changliu and Tomizuka, Masayoshi},
  booktitle={Dynamic Systems and Control Conference},
  year={2014}
}

@inproceedings{ames2019control,
  title={Control barrier functions: Theory and applications},
  author={Ames, Aaron D and Coogan, Samuel and Egerstedt, Magnus and Notomista, Gennaro and Sreenath, Koushil and Tabuada, Paulo},
  booktitle={European control conference},
  year={2019},
}

@article{tayal2024collisionconeapproachcontrol,
      title={A Collision Cone Approach for Control Barrier Functions}, 
      author={Manan Tayal and Bhavya Giri Goswami and Karthik Rajgopal and Rajpal Singh and Tejas Rao and Jishnu Keshavan and Pushpak Jagtap and Shishir Kolathaya},
      year={2024},
      eprint={2403.07043},
      archivePrefix={arXiv},
      primaryClass={cs.RO},
}

@article{thontepu2022control,
  title={Control Barrier Functions in UGVs for Kinematic Obstacle Avoidance: A Collision Cone Approach},
  author={Thontepu, Phani and Goswami, Bhavya Giri and Singh, Neelaksh and PI, Shyamsundar and Sundaram, Suresh and Katewa, Vaibhav and others},
  journal={arXiv preprint arXiv:2209.11524},
  year={2022}
}

@article{tayal2023control,
  title={Control Barrier Functions in Dynamic UAVs for Kinematic Obstacle Avoidance: A Collision Cone Approach},
  author={Tayal, Manan and Kolathaya, Shishir},
  journal={arXiv preprint arXiv:2303.15871},
  year={2023}
}

@article{xiao2021adaptive,
  title={Adaptive control barrier functions},
  author={Xiao, Wei and Belta, Calin and Cassandras, Christos G},
  journal={IEEE Transactions on Automatic Control},
  volume={67},
  number={5},
  pages={2267--2281},
  year={2021},
  publisher={IEEE}
}

@inproceedings{lopez2023unmatched,
  title={Unmatched control barrier functions: Certainty equivalence adaptive safety},
  author={Lopez, B. T. and Slotine, J.-. E.},
  booktitle={American Control Conference},
  pages={3662--3668},
  year={2023},
  organization={IEEE}
}

@INPROCEEDINGS{siaGo2025yun,
  author={Yun, Kai S. and Chen, Rui and Dunaway, Chase and Dolan, John M. and Liu, Changliu},
  booktitle={2025 IEEE International Conference on Robotics and Automation (ICRA)}, 
  title={Safe Control of Quadruped in Varying Dynamics via Safety Index Adaptation}, 
  year={2025},
  volume={},
  number={},
  pages={7771-7777}
}

@article{Cosner2022SelfSupervisedOL,
  title={Self-Supervised Online Learning for Safety-Critical Control using Stereo Vision},
  author={Ryan K. Cosner and Ivan Dario Jimenez Rodriguez and Tam{\'a}s G. Moln{\'a}r and Wyatt Ubellacker and Yisong Yue and A. Ames and Katherine L. Bouman},
  journal={2022 International Conference on Robotics and Automation (ICRA)},
  year={2022},
  pages={11487-11493},
}

@inproceedings{domain-randomization,
author = {Tobin, Josh and Fong, Rachel and Ray, Alex and Schneider, Jonas and Zaremba, Wojciech and Abbeel, Pieter},
title = {Domain randomization for transferring deep neural networks from simulation to the real world},
year = {2017},
publisher = {IEEE Press},
booktitle = {2017 IEEE/RSJ International Conference on Intelligent Robots and Systems (IROS)},
pages = {23–30},
numpages = {8},
}

@inproceedings{hendrycks*2020augmix,
    title={AugMix: A Simple Method to Improve Robustness and Uncertainty under Data Shift},
    author={Dan Hendrycks and Norman Mu and Ekin Dogus Cubuk and Barret Zoph and Justin Gilmer and Balaji Lakshminarayanan},
    booktitle={International Conference on Learning Representations},
    year={2020},
}

@article{MacKay1992APB,
  title={A Practical Bayesian Framework for Backpropagation Networks},
  author={David John Cameron MacKay},
  journal={Neural Computation},
  year={1992},
  volume={4},
  pages={448-472},
}

@article{ames2017cbf,
  author={Ames, Aaron D. and Xu, Xiangru and Grizzle, Jessy W. and Tabuada, Paulo},
  journal={IEEE Transactions on Automatic Control}, 
  title={Control Barrier Function Based Quadratic Programs for Safety Critical Systems}, 
  year={2017},
  volume={62},
  number={8},
  pages={3861-3876},
}

@inproceedings{majumdar2018pacbayes,
  title = 	 {PAC-Bayes Control: Synthesizing Controllers that Provably Generalize to Novel Environments},
  author =       {Majumdar, Anirudha and Goldstein, Maxwell},
  booktitle = 	 {Proceedings of The 2nd Conference on Robot Learning},
  pages = 	 {293--305},
  year = 	 {2018},
  volume = 	 {87},
  series = 	 {Proceedings of Machine Learning Research},
  month = 	 {29--31 Oct},
  publisher =    {PMLR},
}

@misc{contreras2025sodaMPC,
      title={Safe, Out-of-Distribution-Adaptive MPC with Conformalized Neural Network Ensembles}, 
      author={Jose Leopoldo Contreras and Ola Shorinwa and Mac Schwager},
      year={2025},
      eprint={2406.02436},
      archivePrefix={arXiv},
      primaryClass={cs.RO},
}

@misc{seo2025unisafe,
      title={Uncertainty-aware Latent Safety Filters for Avoiding Out-of-Distribution Failures}, 
      author={Junwon Seo and Kensuke Nakamura and Andrea Bajcsy},
      year={2025},
      eprint={2505.00779},
      archivePrefix={arXiv},
      primaryClass={cs.RO},
}

@inproceedings{castaneda2023idCBF,
  title = 	 {In-Distribution Barrier Functions: Self-Supervised Policy Filters that Avoid Out-of-Distribution States},
  author =       {Casta\~neda, Fernando and Nishimura, Haruki and McAllister, Rowan Thomas and Sreenath, Koushil and Gaidon, Adrien},
  booktitle = 	 {Proceedings of The 5th Annual Learning for Dynamics and Control Conference},
  pages = 	 {286--299},
  year = 	 {2023},
  volume = 	 {211},
  series = 	 {Proceedings of Machine Learning Research},
  publisher =    {PMLR},
}

@inproceedings{richter2017safevisualnav, 
    AUTHOR    = {Charles Richter AND Nicholas Roy}, 
    TITLE     = {Safe Visual Navigation via Deep Learning and Novelty Detection}, 
    BOOKTITLE = {Proceedings of Robotics: Science and Systems}, 
    YEAR      = {2017}, 
    ADDRESS   = {Cambridge, Massachusetts}, 
    MONTH     = {July}, 
}

@inproceedings{reichlin2022backtomanifold,
  author={Reichlin, Alfredo and Marchetti, Giovanni Luca and Yin, Hang and Ghadirzadeh, Ali and Kragic, Danica},
  booktitle={2022 IEEE/RSJ International Conference on Intelligent Robots and Systems (IROS)}, 
  title={Back to the Manifold: Recovering from Out-of-Distribution States}, 
  year={2022},
  pages={8660-8666},
}

@article{wellhausen2020anomalyDetection,
  author={Wellhausen, Lorenz and Ranftl, René and Hutter, Marco},
  journal={IEEE Robotics and Automation Letters}, 
  title={Safe Robot Navigation Via Multi-Modal Anomaly Detection}, 
  year={2020},
  volume={5},
  number={2},
  pages={1326-1333},
}

@misc{bansal2024filterdeployallrobust,
      title={One Filter to Deploy Them All: Robust Safety for Quadrupedal Navigation in Unknown Environments}, 
      author={Albert Lin and Shuang Peng and Somil Bansal},
      year={2024},
      eprint={2412.09989},
      archivePrefix={arXiv},
      primaryClass={cs.RO},
}

@misc{chakraborty2024enhancingsafetyrobustnessvisionbased,
      title={Enhancing Safety and Robustness of Vision-Based Controllers via Reachability Analysis}, 
      author={Kaustav Chakraborty and Aryaman Gupta and Somil Bansal},
      year={2024},
      eprint={2410.21736},
      archivePrefix={arXiv},
      primaryClass={cs.RO},
}

@inproceedings{AgileButSafe,
  author    = {He, Tairan and Zhang, Chong and Xiao, Wenli and He, Guanqi and Liu, Changliu and Shi, Guanya},
  title     = {Agile But Safe: Learning Collision-Free High-Speed Legged Locomotion},
  booktitle = {Robotics: Science and Systems (RSS)},
  year      = {2024},
}

@inproceedings{antonio2024conformalPolicyLearning,
  author={Huang, Huang and Sharma, Satvik and Loquercio, Antonio and Angelopoulos, Anastasios and Goldberg, Ken and Malik, Jitendra},
  booktitle={2024 IEEE International Conference on Robotics and Automation (ICRA)}, 
  title={Conformal Policy Learning for Sensorimotor Control under Distribution Shifts}, 
  year={2024},
  pages={16285-16291},
}

@inproceedings{dean2021certaintyequiv,
  title = 	 {Certainty Equivalent Perception-Based Control},
  author =       {Dean, Sarah and Recht, Benjamin},
  booktitle = 	 {Proceedings of the 3rd Conference on Learning for Dynamics and Control},
  pages = 	 {399--411},
  year = 	 {2021},
  volume = 	 {144},
  series = 	 {Proceedings of Machine Learning Research},
  month = 	 {07 -- 08 June},
  publisher =    {PMLR},
}

@inproceedings{wei2025modelverification,
    author="Wei, Tianhao and Hu, Hanjiang and Marzari, Luca and Yun, Kai S. and Niu, Peizhi and Luo, Xusheng and Liu, Changliu",
    title="ModelVerification.jl: A Comprehensive Toolbox for Formally Verifying Deep Neural Networks",
    booktitle="Computer Aided Verification",
    year="2025",
    publisher="Springer Nature Switzerland",
    pages="395--408",
}

@article{mitra2022verify,
  author={Hsieh, Chiao and Li, Yangge and Sun, Dawei and Joshi, Keyur and Misailovic, Sasa and Mitra, Sayan},
  journal={IEEE Transactions on Computer-Aided Design of Integrated Circuits and Systems}, 
  title={Verifying Controllers With Vision-Based Perception Using Safe Approximate Abstractions}, 
  year={2022},
  volume={41},
  number={11},
  pages={4205-4216},
}

@article{dawson2022certificate,
  author={Dawson, Charles and Lowenkamp, Bethany and Goff, Dylan and Fan, Chuchu},
  journal={IEEE Robotics and Automation Letters}, 
  title={Learning Safe, Generalizable Perception-Based Hybrid Control With Certificates}, 
  year={2022},
  volume={7},
  number={2},
  pages={1904-1911},
}

@article{majumdar2021pacbayes,
    author = {Anirudha Majumdar and Alec Farid and Anoopkumar Sonar},
    title ={PAC-Bayes control: learning policies that provably generalize to novel environments},
    journal = {The International Journal of Robotics Research},
    volume = {40},
    number = {2-3},
    pages = {574-593},
    year = {2021},
}

@inproceedings{ryan2021mrcbf,
  author={Cosner, Ryan K. and Singletary, Andrew W. and Taylor, Andrew J. and Molnar, Tamas G. and Bouman, Katherine L. and Ames, Aaron D.},
  booktitle={2021 IEEE/RSJ International Conference on Intelligent Robots and Systems (IROS)}, 
  title={Measurement-Robust Control Barrier Functions: Certainty in Safety with Uncertainty in State}, 
  year={2021},
  pages={6286-6291},
}

@inproceedings{bena2025poisson,
  author={Bena, Ryan M. and Bahati, Gilbert and Werner, Blake and Cosner, Ryan K. and Yang, Lizhi and Ames, Aaron D.},
  booktitle={2025 IEEE-RAS 24th International Conference on Humanoid Robots (Humanoids)}, 
  title={Geometry-Aware Predictive Safety Filters on Humanoids: From Poisson Safety Functions to CBF Constrained MPC}, 
  year={2025},
  pages={1-8},
}

@inproceedings{pmlr-v242-toufighi24a,
  title = 	 {Decision boundary learning for safe vision-based navigation via {H}amilton-{J}acobi reachability analysis and support vector machine},
  author =       {Toufighi, Tara and Bui, Minh and Shrestha, Rakesh and Chen, Mo},
  booktitle = 	 {Proceedings of the 6th Annual Learning for Dynamics and Control Conference},
  pages = 	 {440--452},
  year = 	 {2024},
  volume = 	 {242},
  series = 	 {Proceedings of Machine Learning Research},
  month = 	 {15--17 Jul},
  publisher =    {PMLR},
}

@inproceedings{brown2022unified,
  title={A unified view of SDP-based neural network verification through completely positive programming},
  author={Brown, Robin A and Schmerling, Edward and Azizan, Navid and Pavone, Marco},
  booktitle={International conference on artificial intelligence and statistics},
  pages={9334--9355},
  year={2022},
  organization={PMLR}
}

\newpage
\appendix

\section{Background on Adaptive Perception Error Margin}\label{app: base_error_ratio}

This section complements Sec.~\ref{subsec: adapt_bound} by discussing how the normalized score function used for \textit{conformalizing scalar uncertainty estimates} motivated our construction of the base error ratio, $\varphi_{\text{cal}}$, and the OoD-aware adaptive perception error margin, $\EpsilonAdapt(y)$.
Here, we assume that readers have an understanding of conformal prediction (CP), and refer readers to~\citet{cp_bates_intro} for a comprehensive introduction to CP.

\subsection{Conformalizing Scalar Uncertainty Estimates}\label{app: csue}
As discussed in~\citet{cp_bates_intro}, this procedure creates adaptive prediction sets for regression tasks. In short, an adaptive set scales its width to match the model's confidence, producing narrow, precise sets for inputs it finds ``easy'' and wider, more cautious sets for inputs it finds ``hard.'' The method is as follows\footnote{This methodology is largely based on the work of~\citet{cp_bates_intro}, and the formulation closely follow their presentation.}:

\begin{enumerate}[leftmargin=*]
    \item \textbf{Assume a scalar uncertainty}. We start with a pre-trained regression model $\hat q(y):\mathbb R^l\rightarrow\mathbb R$, where the input is $y\in\mathbb R^l$, prediction output is $\hat x\in\mathbb R$, and true label is $x\in\mathbb R$, such that $\hat x = \hat q(y)$. We also have a separate function $u(y):\mathbb R^l\rightarrow\mathbb R_+$ that produces a scalar uncertainty metric. This $u(y)$ can be any heuristic, such as an estimate of the standard deviation or, in our case, the value from an epistemic uncertainty quantification module $\Unc(y)$. It is designed such that larger values of $u(y)$ encode more uncertainty.
    \item \textbf{Define a normalized score function}. A non-conformity score $s(y,x)$ is defined by normalizing the model's prediction error by this scalar uncertainty:
        \begin{align}\label{eq: norm_score}
            s(y,x) \triangleq \frac{|\hat q(y) -x|}{u(y)}.
        \end{align}
    \item \textbf{Calibrate the quantile}. Using a standard split-conformal process, this score is computed for all $N$-samples in a held-out calibration set. A quantile, $\varphi$, is then found, typically as the $\frac{\lceil (N+1)(1-\alpha) \rceil}{N}$ quantile of these calibration scores, for a desired coverage level $1-\alpha$.
    \item \textbf{Form the adaptive prediction set}. For any new test input $Y_{\text{test}}$, the final prediction set $\mathcal P(Y_{\text{test}})$ is given by:
        \begin{align}\label{eq: adapt_set}
            \mathcal P(Y_{\text{test}}) \triangleq \left[ \hat q(Y_{\text{test}}) - u(Y_{\text{test}})\varphi, \hat q(Y_{\text{test}}) + u(Y_{\text{test}})\varphi \right].
        \end{align}
    This set is inherently adaptive, as its width scales directly with the measured uncertainty $u(Y_{\text{test}})$ for new input.
\end{enumerate}
It is critical to note that the formal guarantee $\mathbb P\left[X_{\text{test}}\in\mathcal P(Y_{\text{test}})\right]\ge 1-\alpha$ only holds if the test data is drawn from the same distribution as the calibration data, and falls under OoD test data.

\subsection{From Normalized Conformal Score Function to the Base Error Ratio}
The normalized score function (\ref{eq: norm_score}) provides the direct motivation for our element-wise base error ratio, $\varphi_{\text{cal}}$, in (\ref{eq: base_error_ratio}). However, we deliberately deviate from the standard procedure of conformalizing scalar uncertainty estimates for two critical reasons related to our problem setting:
\begin{enumerate}[leftmargin=*]
    \item \textbf{Absence of Guarantees under OoD}. The formal statistical guarantee of CP explicitly relies on the test data being drawn from the same distribution as the calibration data. Our problem is fundamentally an out-of-distribution one. Thus, computing a quantile for a desired coverage $1-\alpha$ is not meaningful, as any guarantee derived from it would be invalid upon encountering OoD measurements.
    \item \textbf{Uncertainty Type}. As noted by~\citet{cp_bates_intro}, there is no evidence to believe that a scalar uncertainty score would be directly related to the quantiles of the label distribution, especially an epistemic uncertainty score $\Unc(y)$.
\end{enumerate}

Instead of seeking a statistical quantile for ID coverage, our goal is to find a \textit{worst-case robustness bound} derived from the calibration data. We achieve this by using the $\max$ operation in (\ref{eq: base_error_ratio}). This approach is conceptually equivalent to Step 3 in Appendix~\ref{app: csue}, as mentioned in Sec.~\ref{subsec: adapt_bound}, which seeks to find the single worst-case normalized score function observed in the calibration data.

A significant practical challenge with this $\max$ operation is its extreme sensitivity to statistical outliers. The calibration set $\mathcal D_{\text{cal}}$ may contain outlier data points $(y_i,x_i)$ that exhibit an anomalously low uncertainty score ($\Unc(y_i)\approx 0$) but a non-trivial perception error. Such points would cause the ratio in (\ref{eq: base_error_ratio}) to become arbitrarily large, resulting in a $\varphi_{\text{cal}}$ that renders the downstream safety filter (\ref{eq: atom_op}) overly conservative and unusable. 

Therefore, we introduce the filtering procedure (\ref{eq: unc_filtered_set}) as a crucial stabilization heuristic. The filter hyperparameter $\gamma$ is a tunable parameter that allows a user to adjust the desired level of conservatism by excluding these statistical outliers. The necessity of this step is empirically validated in the ablation study in Sec.~\ref{subsec: exp_2d} (Table~\ref{tab: abl_f1}), which demonstrates that as $\gamma$ increases to include these outliers, the filter's task performance, i.e., reach rate, collapses due to this exact over-conservatism.

Finally, our final adaptive perception error margin, $\EpsilonAdapt(y)=\| \varphi_{\text{cal}}\cdot \Unc(y)\|_2$, is constructed in a way that is analogous to the adaptive prediction set's width, $u(Y_{\text{test}})\varphi$, from (\ref{eq: adapt_set}). We scale a calibrated, worst-case base ratio, $\varphi_{\text{cal}}$, by the real-time uncertainty score, $\Unc(y)$, to determine the final, adaptive margin. The final L2-norm is applied to this element-wise vector to produce a scalar margin $\EpsilonAdapt(y)$. This is done specifically to match the assumptions of the MR-CBF framework in~\citet{mrcbf}, which requires a scalar error bound $\epsilon(y)$ such that $\|e(x)\|_2\le \epsilon(y)$.

\section{Details on Epistemic Uncertainty Quantification Modules}\label{app: euq}
\begin{figure}[htp!]
    \centering
    \includegraphics[width=1.0\linewidth]{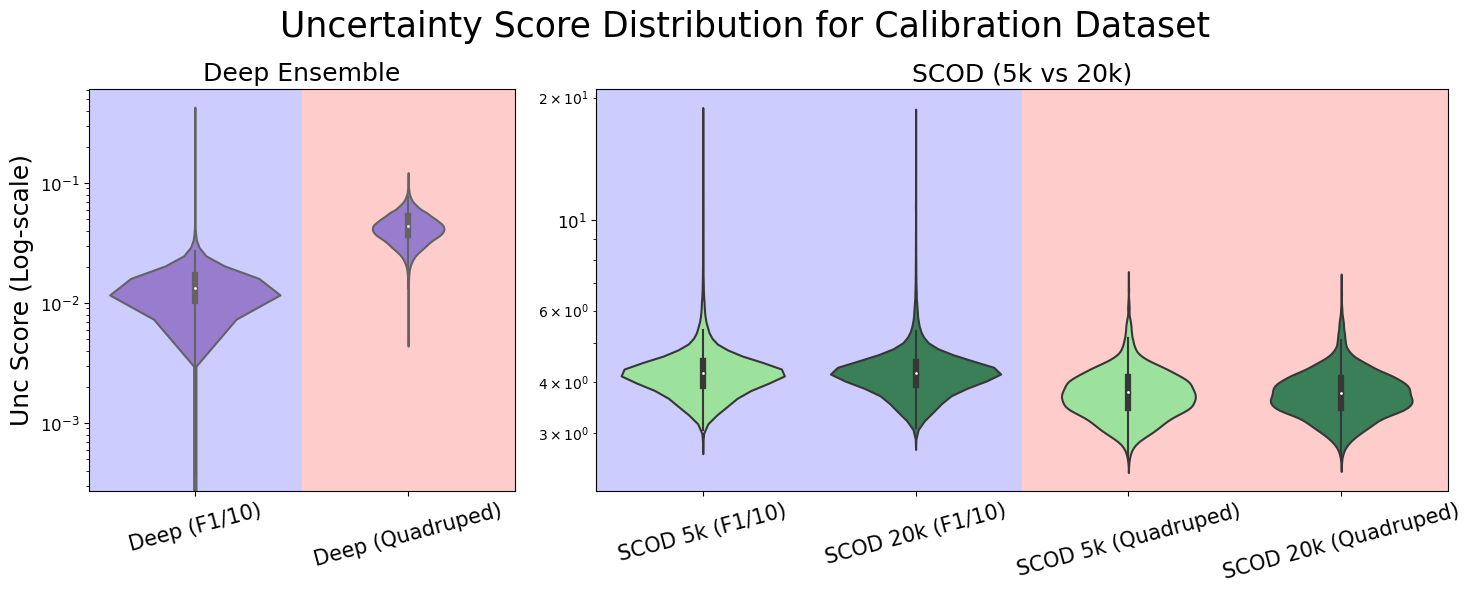}
    \caption{Violin plots comparing the epistemic uncertainty score distributions from the calibration dataset ($\mathcal{D}_{\text{cal}}$), i.e., $S_{\text{cal}}$, for each EUQ module, plotted on a log-scale. The left plot shows the distributions for Deep Ensemble, and the right plot compares SCOD using a 5,000-point sketch (``SCOD 5k'') versus a 20,000-point sketch (``SCOD 20k''). Each plot further separates the distributions for the F1Tenth (light-blue background) and quadruped (light-red background) experiments.}
    \label{fig: cal_unc_comp}
\end{figure}

Here, we provide the configuration details for the EUQ modules. While the underlying perception models for the F1Tenth and quadruped experiments are distinct, the hyperparameters used to build their respective EUQ modules (e.g., number of models for Deep Ensembles, sketch budget for SCOD) were identical for both. 
\begin{itemize}[leftmargin=*]
    \item \textbf{Deep Ensemble}. For both experiments, 5 perception models were trained with distinct random initializations. The epistemic uncertainty is computed as follows, where $M=5$:
    \begin{align}
        \text{Unc}_{\text{Deep}}(y) = \frac1M \left(\sum_{m=1}^M\hat q_m(y)^\top \hat q_m(y)\right) - \mu_*(y)^\top \mu_*(y),
    \end{align}
    where $\hat q_m(y)\in\mathbb R^n$ is the predicted state vector of the $m$-th network and $\mu_*(y)\in\mathbb R^n$ is the element-wise average of the M prediction vectors. 
    \item \textbf{SCOD}. For a given pre-trained perception model, we perform the offline sketching step on a 20,000-point subset of the training data to generate the low-rank approximation of the Fisher matrix. The sketching budget $T$, i.e., the memory budget, was set to 304, the approximation rank $k$ was set to 50, and we used the Subsampled Randomized Fourier Transform (SRFT) sketching operator, as recommended by~\citet{SCOD} for efficiency. To test the sensitivity to the data size, we also performed this same sketching procedure using a smaller 5,000-point subset and observed no significant difference in the quality or magnitude of $\Unc$ values (Fig.~\ref{fig: cal_unc_comp}).
\end{itemize}

Fig.~\ref{fig: cal_unc_comp} plots the distributions of the calibration uncertainty scores, i.e., $S_{\text{cal}}$. The plot visually confirms our ablation study: the distribution for SCOD with 5,000-point subset and 20,000-point subset (right panel) are nearly identical for both experiments. The figure also highlights the vast difference in output scales between the EUQ methods. Deep Ensemble scores (left panel) are on the order of $10^{-2}$, while SCOD scores (right panel) are on the order of $10^0$.

\section{Details on OoD Experiments}\label{app: ood}

\subsection{Details on Safety Filter and Controller}
\textbf{Relaxed ATOM-CBF Filter}. Below is the relaxed variant of the optimization-based ATOM-CBF safety filter use in the experiments:
\begin{align}\label{eq: r_atom_op}
    u_{\text{safe}}(\hat x, \epsilon_{\text{adapt}}(y)) = &\argmin_{u\in\mathcal U} \quad \frac12 \| u - u_{\text{nom}}\|_2^2 + p\delta^2 \\
    \text{s.t.}\quad&\ \mathcal L_fh(\hat x) + \mathcal L_gh(\hat x)u - \epsilon_{\text{adapt}}(y) \big(\mathbb L_{\mathcal L_fh} + \mathbb L_{\kappa\circ h} + \mathbb L_{\mathcal L_gh}\|u\|_2\big) \ge -\kappa(h(\hat x)) -\delta, \nonumber 
\end{align}
where $p\in\mathbb R_+$ is a large coefficient to penalize the slack variable, $\delta$, and $\mathcal U$ is the control limits. Note that although the formal definition of ATOM-CBF (Def.~\ref{def: atom_cbf}) does not account for control limits, we implement them in the experiments for realism.
\\
\\
\noindent\textbf{Controller Parameters}. The extended class $\mathcal K_\infty$ function $\kappa$ is a scalar function in our experiments.
The Lipschitz constants $\mathbb L_{\mathcal L_fh}, \mathbb L_{\mathcal L_gh}, \mathbb L_{\kappa\circ h}$ were estimated by sampling on a set of gridded values on the system's safe set $\mathcal C$, and taking the largest numerical gradient, for each experiment. 
Note that such sampling method for finding the Lipschitz constants was used by~\citet{mrcbf} and~\citet{ryan2021mrcbf}.
Various controller parameter values are shown in Table~\ref{tab: exp_details}.

\begin{table}[htb!]
    \centering
    \caption{Controller parameter values for OoD experiments. $v$ [m/s] and $\omega$ [rad/s].}
    \resizebox{\textwidth}{!}{
    \begin{tabular}{lccccccccccc} 
        \toprule
           & \multicolumn{2}{c}{\textbf{Control Limits}} & \multicolumn{5}{c}{\textbf{Relaxed ATOM-CBF}} & \multicolumn{4}{c}{\textbf{Nominal Controller}}\\
           \cmidrule(lr){2-3} \cmidrule(lr){4-8} \cmidrule(lr){9-12}
           \textbf{Exp.} & $[\underline v, \bar v]$ & $[\underline \omega, \bar \omega]$ & $\kappa$ & $\mathbb L_{\mathcal L_fh}$ & $\mathbb L_{\mathcal L_gh}$ & $\mathbb L_{\kappa\circ h}$ & $p$ & $k_{p, \text{dist.}}$ & $k_{d, \text{dist.}}$ & $k_{p, \text{ang.}}$ & $k_{d, \text{ang.}}$\\
        \midrule
        \textbf{F1/10} & $[0.0, 3.0]$ & $[-1.5,1.5]$ & $4.0$ & $0.00$ & $0.40$ & $4.00$ & $100.0$ & $0.8$ & $0.1$ & $2.5$ & $0.1$\\
        \textbf{Quad.} & $[-1.5,1.5]$ & $[-1.5,1.5]$ & $0.1$ & $0.00$ & $1.66$ & $0.10$ & $100.0$ & $0.8$ & $0.1$ & $2.5$ & $0.1$\\
        \bottomrule
    \end{tabular}
    }
    \label{tab: exp_details}
\end{table}

\subsection{Filtering and Calibration Statistics}

The calibration and filtering process described in Section~\ref{subsec: adapt_bound} is crucial for calculating a stable base error ratio, $\varphi_{\text{cal}}$, as discussed in Sec.~\ref{subsec: exp_2d} and Appendix~\ref{app: base_error_ratio}. Table~\ref{tab: cal_stat} provides the full statistics for this procedure for both the F1Tenth and quadruped experiments.

For both experiments, the filtering hyperparameter $\gamma$ was set to the standard deviation, $\sigma_{\text{unc}}$, of the calibration uncertainty scores $S_{\text{cal}}$. Table~\ref{tab: cal_stat} shows the initial calibration set size ($|\mathcal{D}_{\text{cal}}|$), the mean ($\mu_{\text{unc}}$), and the filtering hyperparameter $\gamma$ of the calibration uncertainty scores. The final filtered set size ($|\mathcal{D}_{\text{filtered}}|$) and the resulting base error ratio $\varphi_{\text{cal}}$ for each EUQ method are also presented.

\begin{table}[htp!]
    \centering
    \caption{Calibration and filtering statistics for the F1Tenth and Quadruped experiments ($\gamma=\sigma_{\text{unc}}$).}
    {\small
    \begin{tabular}{llccccc} 
        \toprule
           & \textbf{EUQ} & $|\mathcal D_{\text{cal}}|$ & $\mu_{\text{unc}}$ & $\gamma \;(= \sigma_{\text{unc}})$ & $|\mathcal D_{\text{filtered}}|$ & $\varphi_{\text{cal}} ([d,\alpha])$\\
        \midrule
        \multirow{2}{*}{\textbf{F1Tenth}} & \textbf{SCOD}        & 40,000  & 4.2794 & 0.7250 & 34,076 & [3.690e-2, 1.777e-2] \\
        & \textbf{Deep}     & 40,000 & 0.0145 & 0.0073 & 33,911 & [12.21, 6.566]\\
        \midrule
        \multirow{2}{*}{\textbf{Quadruped}} & \textbf{SCOD} & 10,000 & 5.3180 & 0.7518 & 7,848 & [4.400e-2, 2.603e-2]\\
        & \textbf{Deep} & 10,000 & 0.0477 & 0.0124 & 6,955 & [8.555, 2.023] \\
        \bottomrule
    \end{tabular}
    }
    \label{tab: cal_stat}
\end{table}

\subsection{Experimental Metrics and Discussion}\label{app: add_exp_details}

\begin{figure}[htb!]
    \centering
    \includegraphics[width=0.70\linewidth]{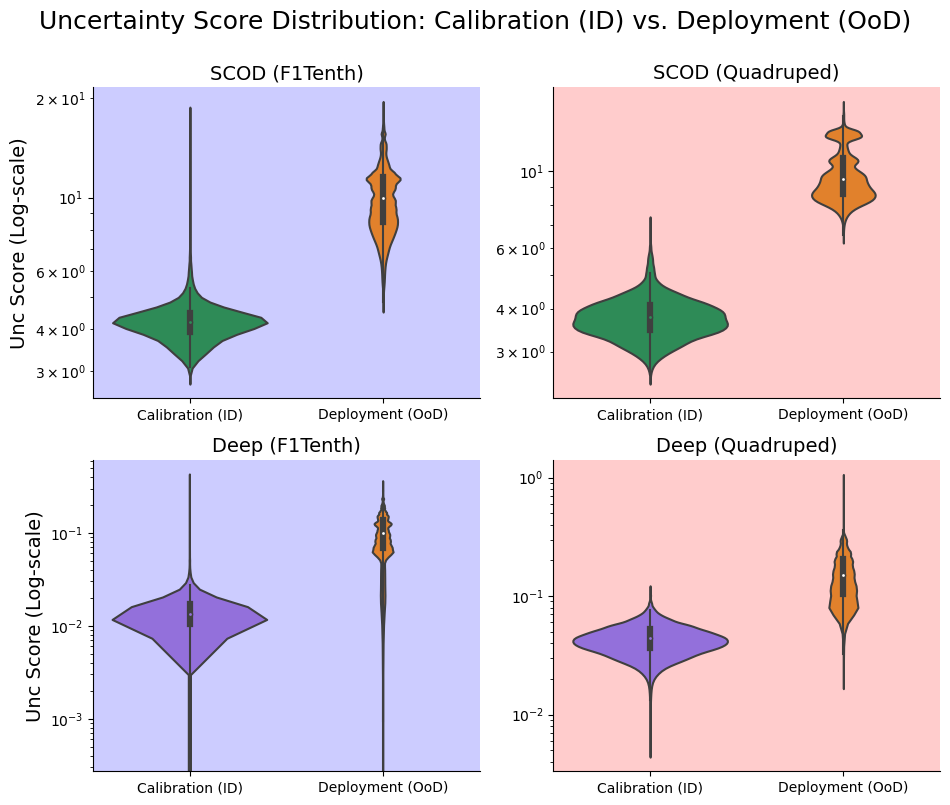}
    \caption{Violin plots comparing the epistemic uncertainty score distributions from the calibration dataset ($\mathcal D_{\text{cal}}$), i.e., $S_{\text{cal}}$, and the deployment OoD measurements. (Top) SCOD, (Bottom) Deep Ensemble, (Left) F1Tenth, (Right) Quadruped. $\Unc$ scores are on a log-scale. Note the magnitudes.}
    \label{fig: cal_vs_ood_unc}
\end{figure}

\begin{figure}[htb!]
    \centering
    \includegraphics[width=1.0\linewidth]{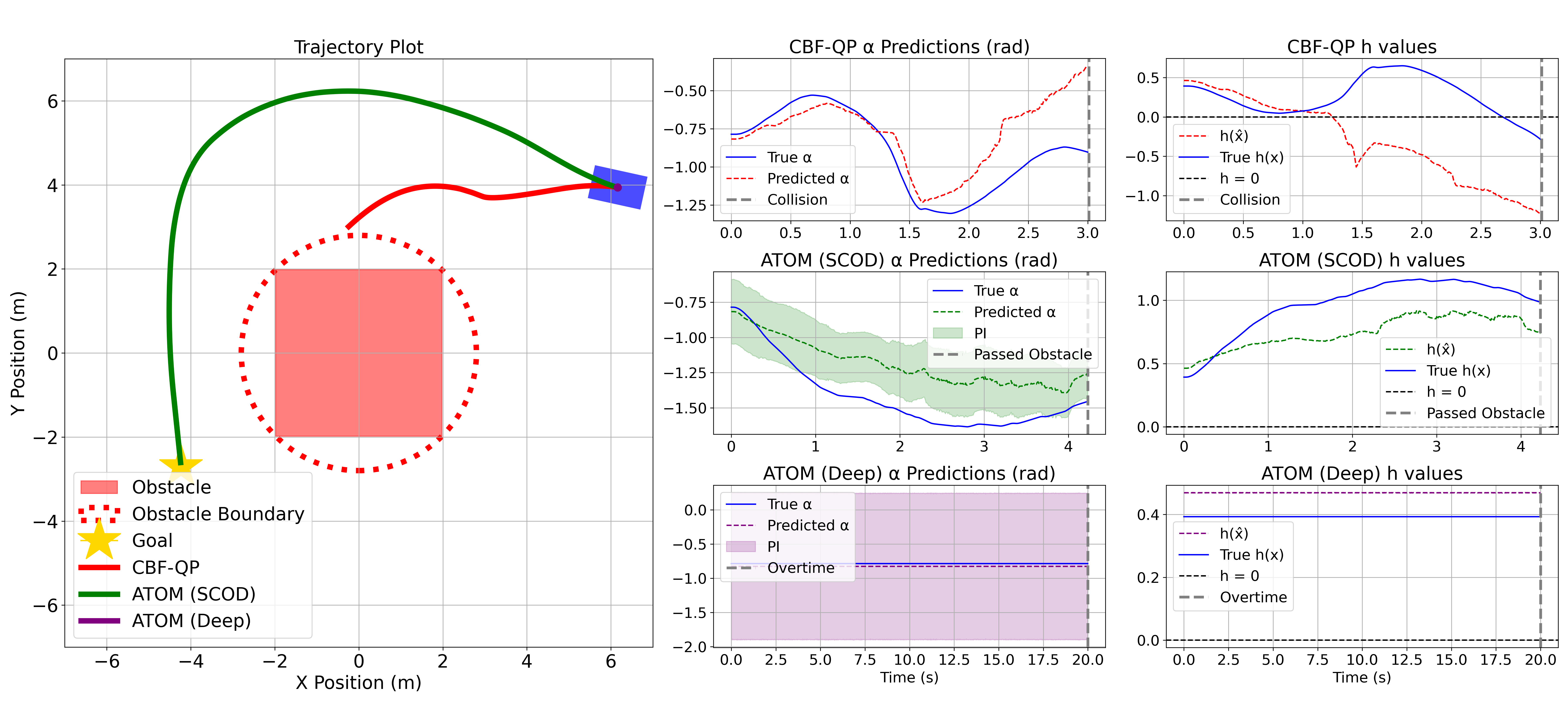}
    \caption{F1Tenth vehicle control experiment with rectangle obstacle, demonstrating over-conservatism in ATOM-CBF (Deep). While ATOM-CBF (SCOD) reaches the goal, ATOM-CBF (Deep) deadlocks at the start, a result of the massive adaptive error bound (visualized by the large purple PI in the bottom-middle plot) generated from high epistemic uncertainty.}
    \label{fig: app_f1_rectangular}
\end{figure}

\textbf{Metrics in Table~\ref{tab: sim_f1}}. For each state component $j\in\{d,\alpha\}$ and at each timestep $t$, we first define an adaptive prediction interval: $\mathcal I_j(t) \triangleq [\hat x_j(t) - \varphi_{\text{cal},j}\cdot \Unc(y(t)), \hat x_j(t) + \varphi_{\text{cal},j}\cdot \Unc(y(t))]$. We then check if the true state $x_j(t)$ falls within this interval. The coverage percentage reported in Table~\ref{tab: sim_f1} is the total number of timesteps where this condition ($x_j(t) \in \mathcal{I}_j(t)$) is true, divided by the total number of timesteps across all 1,000 trajectories. AUROC (area under the receiver-operator characteristic curve) is computed by comparing the $\Unc$ scores from the ID calibration data and the OoD trajectory data.
\\
\\
\noindent\textbf{Over-conservatism of ATOM-CBF (Deep)}.
Fig.~\ref{fig: cal_vs_ood_unc} provides a visual comparison of the epistemic uncertainty score distributions from the ID calibration dataset and the OoD measurements encountered during deployment. These plots, which use a log-scale, highlight the core reason for the difference in safety filter behavior, depending on the choice of EUQ module. For \textbf{SCOD} (top row), the OoD uncertainty scores (orange) are clearly distinguishable and higher than the ID scores (green). However, the magnitude of this shift is moderate; the mean OoD score is roughly 2.5 times larger than the mean ID score. On the other hand, for \textbf{Deep Ensemble} (bottom row), the separation is even more pronounced. The OoD scores (orange) are often an order of magnitude or more larger than the ID scores (purple).
This significant jump in $\Unc$ scores for Deep Ensemble, multiplied with its $\varphi_{\text{cal}}$, creates the massive adaptive error margin $\epsilon_{\text{adapt}}$ that leads to over-conservatism and deadlock shown in Fig.~\ref{fig: app_f1_rectangular}.

\begin{figure}[htb!]
    \centering
    \includegraphics[width=1.0\linewidth]{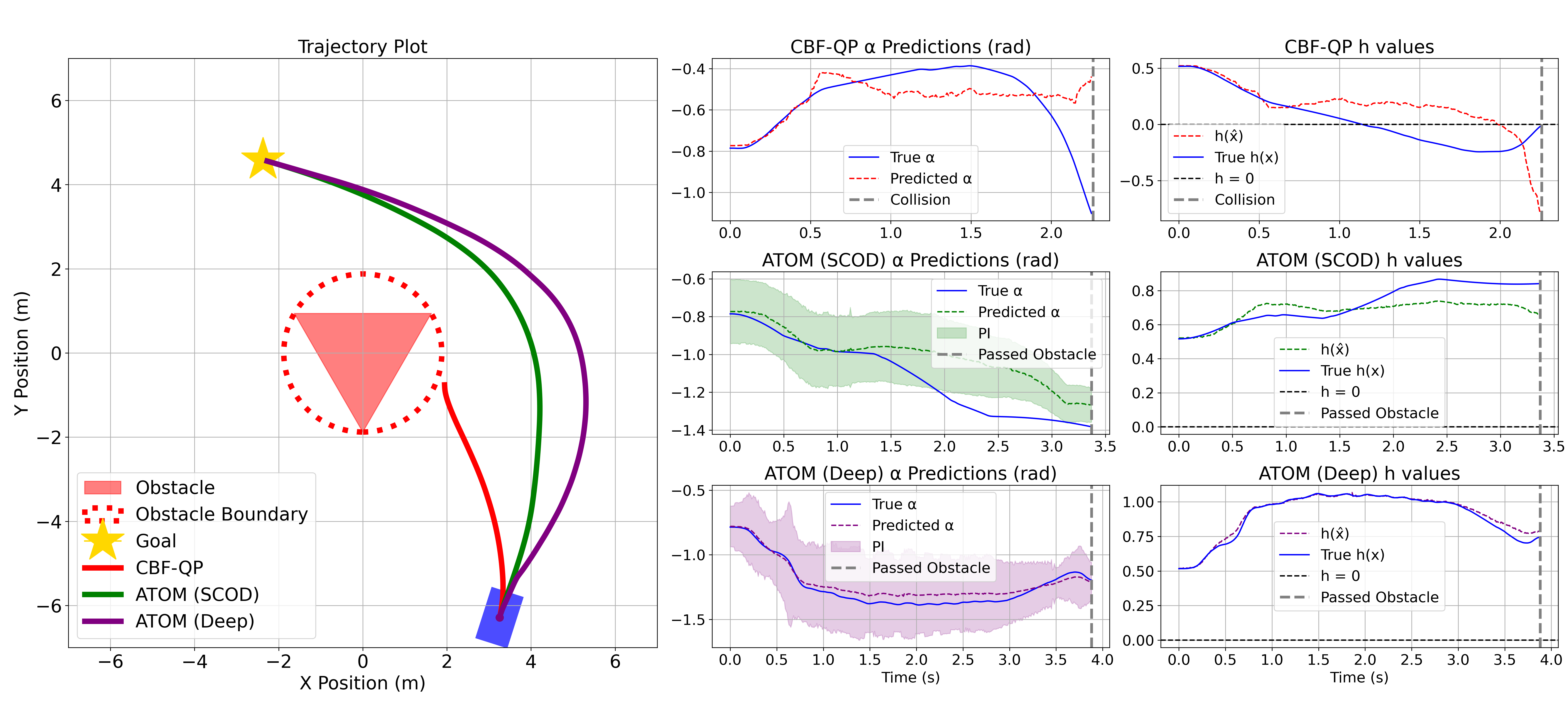}
    \caption{F1Tenth vehicle control experiment with triangle obstacle. Both variants of ATOM-CBF reach the goal without collision, while baseline CBF-QP ends up in a collision.}
    \label{fig: app_f1_triangle}
\end{figure}


\end{document}